\documentclass{article}

\usepackage{microtype}
\usepackage{graphicx}
\usepackage{subfig}
\usepackage{booktabs} 
\usepackage{enumitem}
\setlist[itemize]{leftmargin=*}
\setlist[enumerate]{leftmargin=*}
\usepackage{hyperref}
\usepackage{array}
\newcolumntype{P}[1]{>{\centering\arraybackslash}p{#1}}

\usepackage[accepted]{icml2024}

\usepackage{amsmath}
\usepackage{amssymb}
\usepackage{mathtools}
\usepackage{amsthm}

\usepackage[capitalize,noabbrev]{cleveref}

\theoremstyle{plain}

\theoremstyle{definition}

\theoremstyle{remark}

\usepackage[textsize=tiny]{todonotes}

\usepackage{amsfonts}
\usepackage{textcomp}
\usepackage{xcolor}
\usepackage{comment}
\usepackage{booktabs}
\usepackage{enumitem}
\usepackage{url}
\usepackage[normalem]{ulem}
\usepackage{array}
\usepackage{amsmath}

\icmltitlerunning{Decision Theory-Guided Deep Reinforcement Learning for Fast Learning}

\begin{document}

\twocolumn[
\icmltitle{Decision Theory-Guided Deep Reinforcement Learning for Fast Learning}




\begin{icmlauthorlist}
\icmlauthor{Zelin Wan}{vt}
\icmlauthor{Jin-Hee Cho}{vt}
\icmlauthor{Mu Zhu}{ncsu}
\icmlauthor{Ahmed H. Anwar}{army}
\icmlauthor{Charles Kamhoua}{army}
\icmlauthor{Munindar P. Singh}{ncsu}
\end{icmlauthorlist}

\icmlaffiliation{vt}{Department of Computer Science, Virginia Tech, Falls Church, VA, USA}
\icmlaffiliation{ncsu}{Department of Computer Science, North Carolina State University, Raleigh, NC, USA}
\icmlaffiliation{army}{DEVCOM Army Research Laboratory, Adelphi, MD, USA}

\icmlcorrespondingauthor{Zelin Wan}{zelin@vt.edu}
\icmlcorrespondingauthor{Jin-Hee Cho}{jicho@vt.edu}

\icmlkeywords{Deep reinforcement learning, decision theory, proximal policy optimization, cold start Problem, integration strategies, cart pole challenge, maze navigation}

\vskip 0.3in
]



\printAffiliationsAndNotice{\icmlEqualContribution} 

\begin{abstract}
This paper introduces a novel approach, Decision Theory-guided Deep Reinforcement Learning (DT-guided DRL), to address the inherent cold start problem in DRL. By integrating decision theory principles, DT-guided DRL enhances agents' initial performance and robustness in complex environments, enabling more efficient and reliable convergence during learning. Our investigation encompasses two primary problem contexts: the cart pole and maze navigation challenges. Experimental results demonstrate that the integration of decision theory not only facilitates effective initial guidance for DRL agents but also promotes a more structured and informed exploration strategy, particularly in environments characterized by large and intricate state spaces. The results of experiment demonstrate that DT-guided DRL can provide significantly higher rewards compared to regular DRL. Specifically, during the initial phase of training, the DT-guided DRL yields up to an 184\% increase in accumulated reward. Moreover, even after reaching convergence, it maintains a superior performance, ending with up to 53\% more reward than standard DRL in large maze problems. DT-guided DRL represents an advancement in mitigating a fundamental challenge of DRL by leveraging functions informed by human (designer) knowledge, setting a foundation for further research in this promising interdisciplinary domain.
\end{abstract}

\section{Introduction}
\subsection{Motivation}
{\em Deep Reinforcement Learning} (DRL) offers a potent framework for acquiring intricate behaviors. Yet, it often demands an extensive learning period to achieve the desired performance, limiting its applicability in numerous real-world contexts~\cite{dulac2019challenges}. From the early 2000s to the present, researchers have investigated {\em Transfer Learning} as a potential solution to this challenge~\cite{taylor2009transfer,zhu2023transfer,tessler2017deep}. Despite the promise of transfer learning, its efficacy is often hindered in real-world scenarios due to the unavailability of suitable pre-trained models for transferring knowledge or parameters. Concurrently, there is a growing focus within the DRL research community on improving {\em sample efficiency}~\cite{lin2021juewu,den2022reinforcement}.  However, sample efficiency tends to ignore the safety concerns posed by the unacceptable low performance of RL agents in the initial training phase. Although notable progress has been made in both transfer learning and sample efficiency, DRL agents now effectively utilize state abstractions~\cite{hutsebaut2022hierarchical}, but challenges remain. These challenges include unusable performance during early training stages or the difficulty in obtaining appropriate pre-trained models for effective transfer learning.

Decision theory (DT) offers a structured approach to making choices in situations with uncertain outcomes. Drawing from its foundation in {\em utility theory}~\cite{fishburn1979utility} and probability theory~\cite{renyi2007probability}, decision theory provides clear methodologies to make optimal decisions even when outcomes are not perfectly predictable~\cite{north1968tutorial}. The strength of decision theory lies in its ability to integrate all available data and align it with the decision-maker's preferences to determine the most favorable course of action. However, its performance is constrained in complex environments.

\subsection{Key Objectives}
Our approach, Decision Theory-guided Deep Reinforcement Learning (DT-guided DRL), addresses two major DRL challenges: poor early training performance and reliance on often inaccessible pre-trained models. DT-guided DRL incorporates decision theory to mitigate the 'cold start' issue, improving DRL's early-stage effectiveness. This makes the learning process {\em safer}, avoiding potentially harmful random exploration, and {\em faster}, reducing the overall training time. By doing so, DT-guided DRL ensures more reliable and efficient learning outcomes from the start, making it more applicable to real-world scenarios. This detailed integration of decision theory into DRL provides a nuanced solution to enhance DRL's initial training phase, contributing significantly to its practical deployment.

Our {\bf key contributions} are as follows:
\begin{itemize}
\item {\bf Benchmarking Platforms}: We identify common platforms to evaluate DRL and decision theory performance by integrating them to take benefits of both approaches.

\item {\bf Addressing Cold Start in DRL}: We develop a novel DT-guided DRL approach, which harnesses decision theory to enhance the DRL neural network's output, offering effective heuristics in the initial exploration phase, thereby addressing the cold start problem in DRL.

\item {\bf Experimental Comparisons}: Under various standard simulation scenarios, our DT-guided DRL approach outperformed existing baselines and well-known counterparts, including traditional decision theory~\cite{renyi2007probability}, DRL only~\cite{schulman2017proximal}, transfer learning~\cite{zhu2023transfer}, sample efficiency~\cite{ma2022fresher}, and imitation learning~\cite{gros2020tracking}. This marks a significant advancement in both decision theory and deep learning. 
\end{itemize}

\section{Related Work} \label{sec:rw}
\subsection{Platforms for DRL's Performance Analysis}

Table~\ref{tab:platforms} highlights crucial platforms for evaluating Deep Reinforcement Learning (DRL) and decision theory methodologies. Gymnasium (previously OpenAI Gym) stands out in DRL research, offering various environments, including classic control tasks and Atari 2600 games. Specifically, Gym-Maze, tailored for maze navigation challenges, provides a distinct testbed to assess the adaptability and effectiveness of decision theory approaches and DRL agents in intricate, obstacle-rich scenarios.

\begin{table*}[h]
\caption{Platforms for DRL and Decision Theory Evaluation}
    \centering
    \begin{tabular}{l l l}
        \toprule
        Platform & Description & Reference \\
        \midrule
        Gymnasium (OpenAI Gym) & Varied environments for DRL & \cite{brockman2016openai}, \cite{towers_gymnasium_2023} \\
        Gym-Maze & 2D maze navigation & \cite{chan_gym-maze_2023} \\
        Atari 2600 (ALE) & Classic video game suite & \cite{bellemare2013arcade} \\
        MuJoCo & High-fidelity simulation & \cite{todorov2012mujoco} \\
        DeepMind Lab & 3D learning environments & \cite{beattie2016deepmind} \\
        \bottomrule
    \end{tabular}
    \label{tab:platforms}
\end{table*}

\subsection{Mitigating Cold Start Problems in DRL}

{\em Transfer Learning} (TL) has been explored to address the cold start problem in DRL. Notable approaches include the Hierarchical Deep Reinforcement Learning Network (H-DRLN) for lifelong learning in Minecraft, demonstrating superior performance and efficiency by reusing learned skills across tasks~\cite{tessler2017deep}. Similarly, Reinforcement Learning Building Optimizer with Transfer Learning (ReLBOT) applied trained algorithms from data-rich buildings to new ones, addressing cold start for smart building~\cite{genkin2022using}.  However, the success of transfer learning hinges on choosing a pre-training task closely aligned with the target task, but not all models are suitable for this. Invalid assumptions in knowledge transfer can lead to reduced performance in the target domain.

{\em Imitation Learning}, such as the Reduction-based Active Imitation Learning (RAIL) algorithm, reduces the need for extensive expert queries by applying active i.i.d. learning strategies, showing effectiveness in domains like the Cart-pole environment~\cite{gros2020tracking}. The DDPG-IL algorithm combines DDPG with imitation learning for autonomous driving, achieving faster convergence and performance improvements~\cite{zou2020end}. However, AIL is limited by its reliance on high-quality demonstrations and human expertise, often struggling with complex high-dimensional spaces and generalization to new scenarios. Additionally, ethical and safety concerns, along with the need for sophisticated query algorithms, add complexity to its application in diverse domains.

{\em Warm start methods} facilitate efficient DRL by leveraging prior knowledge. \citet{silva2021encoding} integrated domain-specific rules into RL, demonstrating improved performance without extensive data, showing potential for real-world applications.  \citet{wang2024flexnet} enhanced Hybrid Electric Vehicle (HEV) energy management by initializing neural networks with expert knowledge, significantly boosting learning speed and reducing energy consumption. Further, \citet{xu2020learning} used warm-start Q-learning to HEVs for reducing learning iterations and improving fuel efficiency over traditional methods.  \citet{wexler2022analyzing} developed a method of confidence-constrained learning to mitigate the degradation in Warm-Start RL, offering a balance between policy gradient and constrained learning to optimize performance and minimize degradation.  A warm start in DRL and RL entails initializing a model with weights or policies from related tasks, which can accelerate training. However, this approach may lead to biased exploration, limiting the model's ability to fully explore the state-action space of the new task.

Efforts to enhance {\em sample efficiency} in DRL have seen notable advancements. ``JueWu-MC" employed a hierarchical approach and human demonstrations to tackle partial observability in open-world games, significantly outpacing baselines in competitions such as NeurIPS MineRL~\cite{lin2021juewu}. Option Machines (OMs) leveraged high-level instructions for action selection, showing promise in single-task, multi-task, and zero-shot learning environments~\cite{den2022reinforcement}. Additionally, a DRL method designed for cloud-native Service Function Chains (SFC) caching addressed cold-start problems through graph neural networks (GNNs) to optimize processing latency and request acceptance, showcasing superior performance under high load conditions and in cold-start scenarios~\cite{zhang2022coldstart}.  \citet{ma2022fresher} showed a negative correlation between experience freshness and replay frequency, introducing a {\em freshness discounted factor} $\mu$ in prioritized experience replay (PER) and proposing a novel experience replacement strategy to enhance learning efficiency.
Despite these efforts, sample efficiency in RL is limited by its high data requirements, poor performance in sparse reward environments, challenges in modeling complex real-world dynamics, issues with generalization, dependency on initial conditions, and difficulty balancing exploration and exploitation. These factors make it challenging to apply RL effectively in real-world scenarios with limited data.

\section{Decision Theory-Guided DRL} \label{sec:our-approach}

In DRL, striking a balance between exploration and exploitation is crucial for an agent's efficient search for optimal solutions. The traditional approach, known as dynamic $\epsilon$-greedy exploration, empowers agents to initially make random choices due to limited knowledge and gradually transition to a strategy of exploitation as they acquire more information. However, this technique tends to delay the convergence to near-optimal solutions.

Our DT-guided DRL method combines decision theory (DT) with DRL to guide initial agent decisions, replacing random exploration with informed action distributions. This approach not only mitigates cold start but also lessens the likelihood of the agent getting stuck in local optima, offering a more strategic beginning in the learning process.

Unlike existing solutions to mitigate cold start problems, as discussed in Section~\ref{sec:rw}, DT-guided DRL offers effective initial guidance for DRL agents, avoiding poor performance from random exploration and operating efficiently on small datasets, thereby enhancing generalizability across various RL scenarios.

\subsection{Problem Formulation Using DRL} \label{subsec:ours-drl}

Given the popularity of the Cart Pole environment from Gymnasium, as highlighted in previous works~\cite{hsiao2022unentangled,manrique2020parametric}, along with the frequent consideration of maze problems~\cite{dayan2008decision} in decision theory, we have chosen these examples to illustrate our technique.

We adopt the Proximal Policy Optimization (PPO) algorithm~\cite{schulman2017proximal}. We chose PPO because of its superiority over other DRL algorithms, as observed in our experiments. 

A DRL agent is designed for Markov decision processes (MDP), consisting of the following.
\begin{itemize}
\item State ($S$): This refers to a set of all possible states. For cart pole problems, the states encompass the cart position, cart velocity, pole angle, and pole angular velocity. In maze problems, the states consist of the $x$ and $y$ positions, counting from the entry point. 
\item Action ($A$): This is a set of all possible actions. For cart pole problems, the actions include pushing the cart left or right. In maze problems, the actions refer to moving up, down, left, and right. 
\item Transition Probabilities ($T(s'|s, a)$): This refers to the probability of transitioning from one state ($s$) to another state ($s'$), given an action ($a$).
\item Reward ($R(s)$): This is an immediate reward received after transitioning to the state ($s$). In the cart pole problem, a reward of +1 is given for each step taken. In the maze problem, a small negative value is assigned for each step that fails to reach the exit point, whereas a reward of +1 is given upon reaching the exit point.
\item Policy ($\pi$): It refers to a mapping from states to actions.
\end{itemize}

\subsection{Problem Formulation Using Decision Theory}

To create a decision theory (DT) agent that effectively guides a DRL agent, it is essential to tailor the utility function to suit the specific characteristics of the problem or environment being addressed.

\subsubsection{Cart Pole Environment}
In the cart pole problem, where the objective is to prevent the pole from falling by controlling the cart's movements (pushing left or right), we have designed a utility function focused on the pole's angle. This function assigns a positive utility for pushing left when the angle is negative and, conversely, a positive utility for pushing right when the angle is positive, directly supporting the goal of keeping the pole upright. The utility function is defined as follows:
\begin{equation}
U(s_{\mathrm{pole\_angle}}, a) = 
\begin{cases}
-\frac{s_{\mathrm{pole\_angle}}}{0.209}, & \text{if $a$ is push left}\\
\frac{s_{\mathrm{pole\_angle}}}{0.209}, & \text{otherwise.}    
\end{cases}
\label{eq: cart pole utility}
\end{equation}
The variable $s_{\mathrm{pole\_angle}}$ represents the pole angle, ranging between (-0.418, 0.418) radians in the cart pole game. Given that the game ends if the pole angle falls outside the range of (-0.209, 0.209) radians (or $\pm12^{\circ}$), we normalize this value by dividing by 0.209.

\subsubsection{Maze Environment}
The Maze environment is a grid-based setting where the agent has four movement options: up, down, left, and right. The observation space is defined by the agent's current grid position, represented as $(P_{s,x}, P_{s,y})$. For action selection, we employ a decision theory agent that incorporates the distance from the agent's current position to the exit point within its utility function.
We design the utility function as follows:
\begin{equation}
U(s, a) = 
\begin{cases}
\frac{1}{\mathrm{dis}(P_{s'}, P_{\mathrm{exit}})}, & \text{if no obstacles on $a$'s direction}\\
0, & \text{otherwise,}
\end{cases}
\label{eq:maze utility}
\end{equation}
where $s'$ represents the anticipated new state after taking action $a$, and $dis(P_{s'}, P_{exit})$ denotes the distance from the position in state $s'$ to the exit point, which is given by:
\begin{equation}
\small 
\mathrm{dis}(P_{s'}, P_{\mathrm{exit}}) = \sqrt{(P_{s', x} - P_{\mathrm{exit}, x})^2 + (P_{s', y} - P_{\mathrm{exit}, y})^2}.
\label{eq:distance}
\end{equation}

\subsection{Integrating DT with DRL}
\begin{figure*}[h]
\centering
\includegraphics[width=\textwidth]{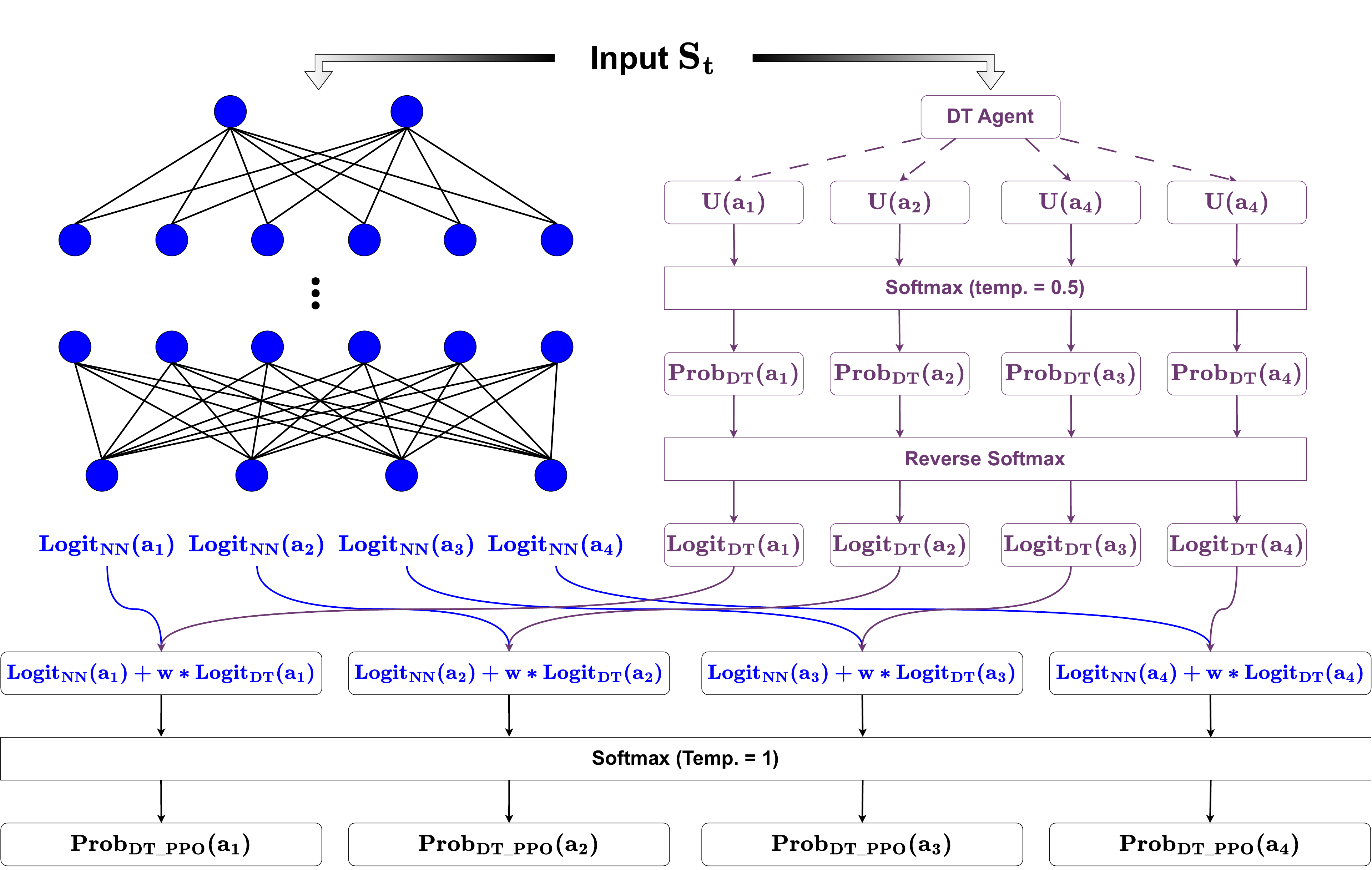}
\caption{{\bf The procedures generating the solutions by a DT-guided DRL agent:} $\mathcal{S}_t$ is the state at round $t$ and $\pi(a_t|\mathcal{S}_t)$ is the probability of all actions. }
\label{fig: DT-guided DRL}
\end{figure*}

The effectiveness of DT-guided DRL lies in its innovative combination of the action probability from the DT agent with the neural network's output. As described in Figure~\ref{fig: DT-guided DRL}, DT-guided DRL operates as follows:
\begin{enumerate}
\item {\bf Utility Function Development}: Create a DT agent with a problem-specific utility function, like Eq.~\eqref{eq: cart pole utility} for the cart pole or Eq.~\eqref{eq:maze utility} for the maze problem. Use a {\em softmax layer} to convert these utility values into a probability distribution for actions, ensuring the values are within the $[0, 1]$ range. A lower temperature setting in the softmax layer increases the DT agent's determinism.

\item {\bf Deep Neural Network (DNN) Model}: Build a DNN that inputs the current state of the environment and outputs probabilities for each possible action.

\item {\bf Integration of DT Agent and Neural Network}: Typically, a softmax function is applied as the final layer in a neural network (NN). In DT-guided DRL, however, we combine the NN's outputs with the DT agent's reverse softmax probabilities prior to the final softmax layer. This involves summing them and dynamically adjusting the weight $w$ applied to the DT agent's probabilities, starting at $w=1$ and gradually decreasing to $w=0$ with each training step.

\item {\bf Final Softmax Layer}: Add a softmax layer with a temperature setting of 1 to the combined outputs, ensuring a valid probability distribution for action selection. A temperature of 1 is standard for NN training.

\item {\bf Neural Network Training}: Train the DNN using a RL algorithm like PPO or DQN. This process includes interacting with the environment, selecting actions based on the combined outputs, receiving rewards, and updating the network's weights to improve action selection.
\end{enumerate}

\section{Experimental Setup}

{\bf Parameterization.} Our study employs two distinct application scenarios: the cart pole and the maze problem. The cart pole scenario, a benchmark in reinforcement learning, challenges the agent to maintain a pole's balance on a cart by applying forces. The maze problem, prevalent in decision theory research, involves an agent navigating through a maze to reach a goal while avoiding obstacles.

For the cart pole, the state space encompasses four continuous variables: cart position and velocity, along with the pole's angle and angular velocity. The action space is binary, involving left or right which indicates the direction of the fixed force the cart is pushed with. The reward is +1 for each timestep the pole remains upright, with the episode ending if the pole tilts beyond $\pm 12^{\circ}$, the cart exits predefined bounds, or after 100 timesteps.

For the maze scenario, the state space is represented by the agent's $x/y$ position within the maze. The agent starts at the top left and aims for the exit at the bottom right. The action space includes four directions: up, down, left, and right. Reaching the exit yields $a+1$ reward, while each step incurs a slight penalty. Episodes conclude upon reaching the exit or surpassing a maximum step count.

In configuring the neural network for our experiments, we adhere to a set of parameters optimized to balance computational efficiency and the capability to learn complex behaviors. Table~\ref{tab:parameters} delineates these parameters, including the discount factor, learning rate, replay buffer size, batch size, clipping parameter, network architecture, and the chosen activation function. These configurations are based on the default settings provided by Stable-Baselines3 \cite{stable-baselines3}, ensuring that our experimental setup is both rigorous and reproducible. The chosen parameters are pivotal in facilitating the agent's learning process, enabling it to effectively navigate and make decisions within the complexities of the cart pole and maze scenarios.

To guarantee the reliability of our results, we base our findings on the average of 100 independent runs for each simulation. This approach reduces the influence of random variations, offering a more dependable assessment of the agent's performance.

\begin{table}[h]
\centering
\caption{Neural Network Configuration Parameters}
\label{tab:parameters}
\vspace{1mm}
\begin{tabular}{l l}
\toprule
\textbf{Parameter}       & \textbf{Value}  \\ \midrule
Discount Factor          & 0.99            \\
Learning Rate            & 0.0003          \\
Replay Buffer Size       & 2048            \\
Batch Size               & 64              \\
Clipping Parameter       & 0.2             \\
Network Architecture     & 64 x 64         \\
Activation Function      & Tanh            \\ \bottomrule
\end{tabular}
\vspace{-2mm}
\end{table}

{\bf Comparing Schemes.} To illustrate our DT-guided DRL approach, we use the Decision Theory-guided Proximal Policy Optimization (DT-guided PPO) as a case study. We compare the DT-guided PPO agent with five alternatives: a pure decision theory-based agent, a standard DRL agent using PPO without DT guidance, a Transfer Learning-enhanced PPO (TL PPO) agent, a sample efficiency-enhanced PPO (SE PPO) agent, and an imitation learning-enhanced PPO (IL PPO). The TL PPO starts with a model pre-trained on a smaller 3$\times$3 maze for 100 episodes and then transitions to a larger maze. The evaluation of TL PPO is confined to maze environments to assess the benefits of transfer learning in spatial tasks. The SE PPO uses experience replay two times more than the standard PPO. Since more frequent experience replay may raise some issues like overfitting and instability, we use the freshness prioritized experience replay~\cite{ma2022fresher}  (an improved alternative version of prioritized experience replay~\cite{schaul2015prioritized}) to mitigate those concerns. The IL PPO leverages an expert policy to generate state-action pairs for 300 episodes, providing a rich set of expert demonstrations for the IL PPO agent. Following this, the agent continues its training in a standard fashion. The expert policy used here is a pre-trained PPO model optimized for maximum accumulated rewards in the cart-pole environment.

This comparative analysis highlights the value of incorporating decision theory into DRL by contrasting it with conventional methods (DT, PPO) and a common approach for addressing cold start issues (TL PPO, SE PPO, and IL PPO). These alternatives serve as benchmarks, enabling a comprehensive evaluation of the DT-guided PPO's performance and effectiveness. Through our graphical representations, which plot episode counts against measures of effectiveness and efficiency, we illustrate the learning dynamics and operational efficiency of each approach, offering a clear comparative perspective.

{\bf Metrics.} Our evaluation relies on two key metrics:(1) {\bf Accumulated reward}: An effectiveness measure, this metric quantifies the agent's ability to maximize returns over time, indicating its task proficiency.  (2) {\bf Running time per step}: Serving as an efficiency metric, it assesses the computational intensity of the decision-making process for each agent, shedding light on the balance between complexity and speed of performance.

\section{Simulation Results \& Analysis}

\subsection{Performance Analysis of the Cart Pole Problem}

\begin{figure}[h]
\centering
\subfloat{\includegraphics[width=0.45\textwidth]{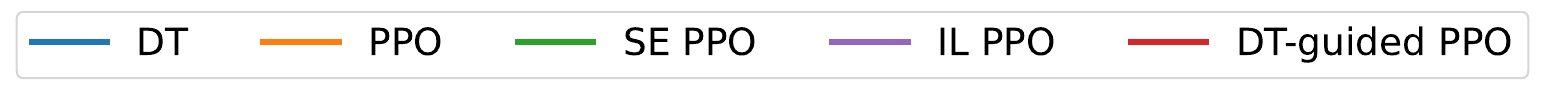}}
\hfil
\vspace{-4mm}
\setcounter{subfigure}{0}
\subfloat{\includegraphics[width=0.45\textwidth]{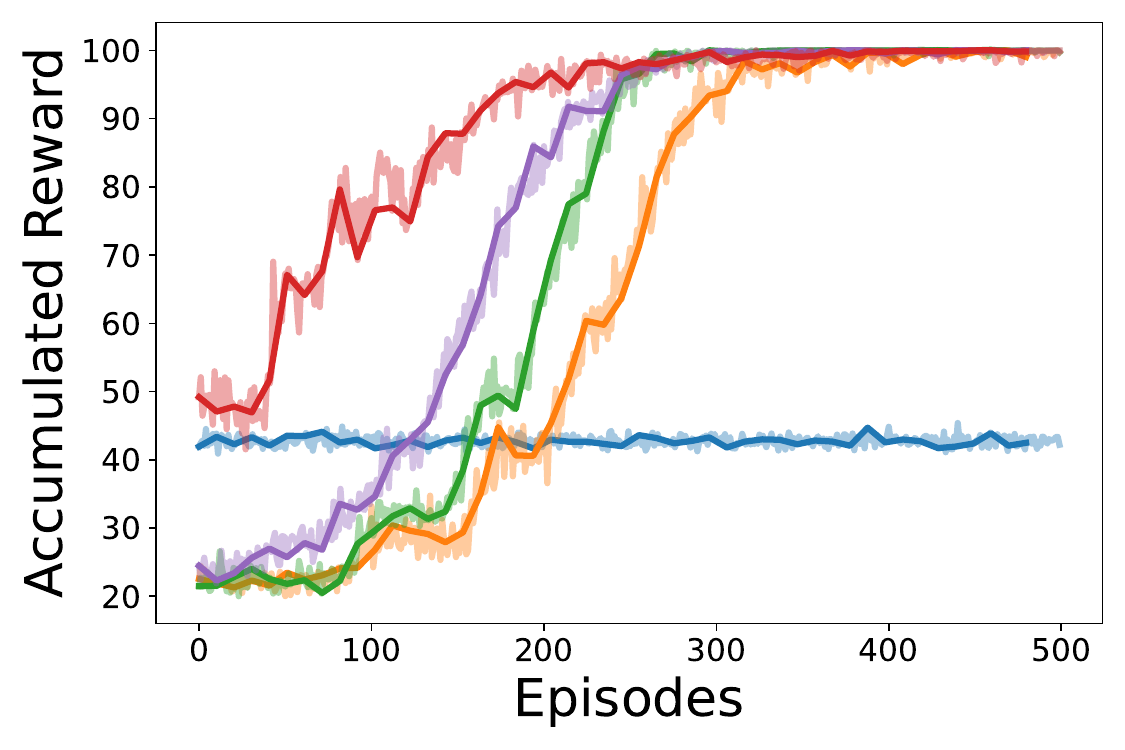} \label{fig: label}}
\hfil
\caption{Comparison of training performance in cart pole problem under DT, PPO, SE PPO, IL PPO, and DT-guided PPO using 500 training episodes in accumulated reward.} \label{fig: cartpole_reward}
\end{figure}

Figure \ref{fig: cartpole_reward} illustrates the accumulated rewards for five agents -- DT, PPO, SE PPO, IL PPO, and DT-guided PPO, within the Cart Pole simulation. We observe the following.

First, the DT-guided PPO agent (red curve) demonstrates superior performance over the other PPO agents, achieving a higher starting reward. This early advantage is attributable to integrating DT utility values with the neural network's output. Given that the neural network begins with zero-initialized outputs post-initialization, and Tanh is employed as the activation function, the utility component predominates in the initial decisions of the DT-guided PPO agent. Consequently, the early behavior of the DT-guided PPO closely mirrors that of the DT agent (blue curve). 

Second, the reward for the DT agent remains static, reflecting its inability to adapt beyond its preconfigured utility function-based decisions. In contrast, the DT-guided PPO agent exhibits continuous enhancement in reward accumulation. This improvement signals the neural network's growing proficiency in recognizing environmental patterns and refining its action strategy accordingly, leading to a progressive increase in accumulated reward. 

Third, as training progresses, both the DT-guided PPO and other PPO agents converge toward similar reward plateaus. However, the DT-guided PPO shows a faster convergence, underscoring the hybrid model's accelerated learning capability. This quicker convergence can be ascribed to the DT-guided PPO agent's initial advantage, which provides a more informed starting point for the learning process and reduces the time spent in exploration.

Figure \ref{fig: cartpole_reward} underpins the assertion that the incorporation of decision theory within the PPO framework yields a dual benefit: mitigating the randomness of actions during the nascent stages of training and facilitating swifter convergence to optimal performance levels in the cart pole challenge.

\begin{figure*}[h]
\centering
\subfloat{\includegraphics[width=0.6\textwidth]{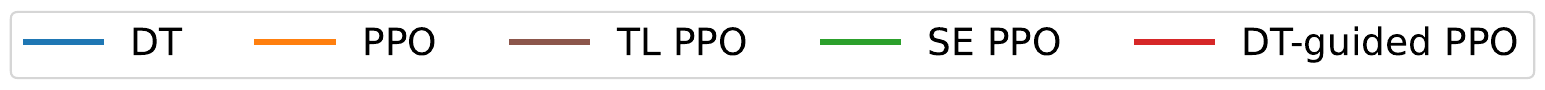}}
\hfil

\setcounter{subfigure}{0}
\subfloat[$m  = 3$]{\includegraphics[width=0.32\textwidth]{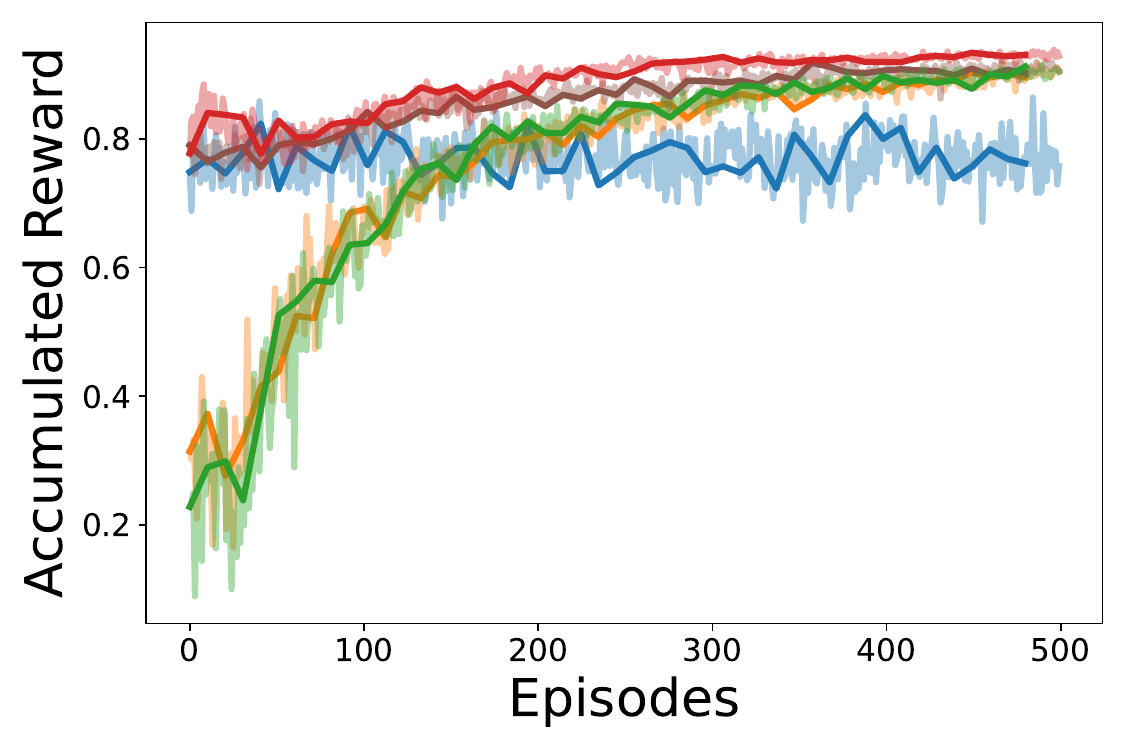} \label{fig:maze_size_3}}
\hfil
\subfloat[$m  = 4$]{\includegraphics[width=0.32\textwidth]{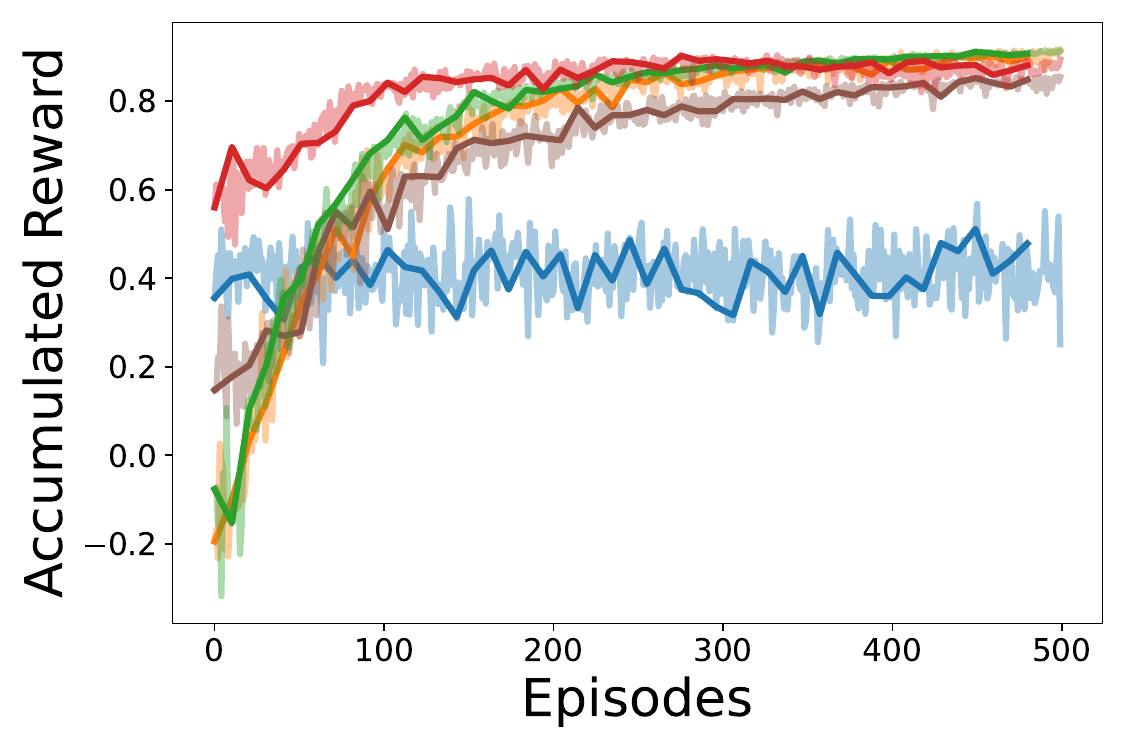} \label{fig:maze_size_4}}
\hfil
\subfloat[$m  = 5$]{\includegraphics[width=0.32\textwidth]{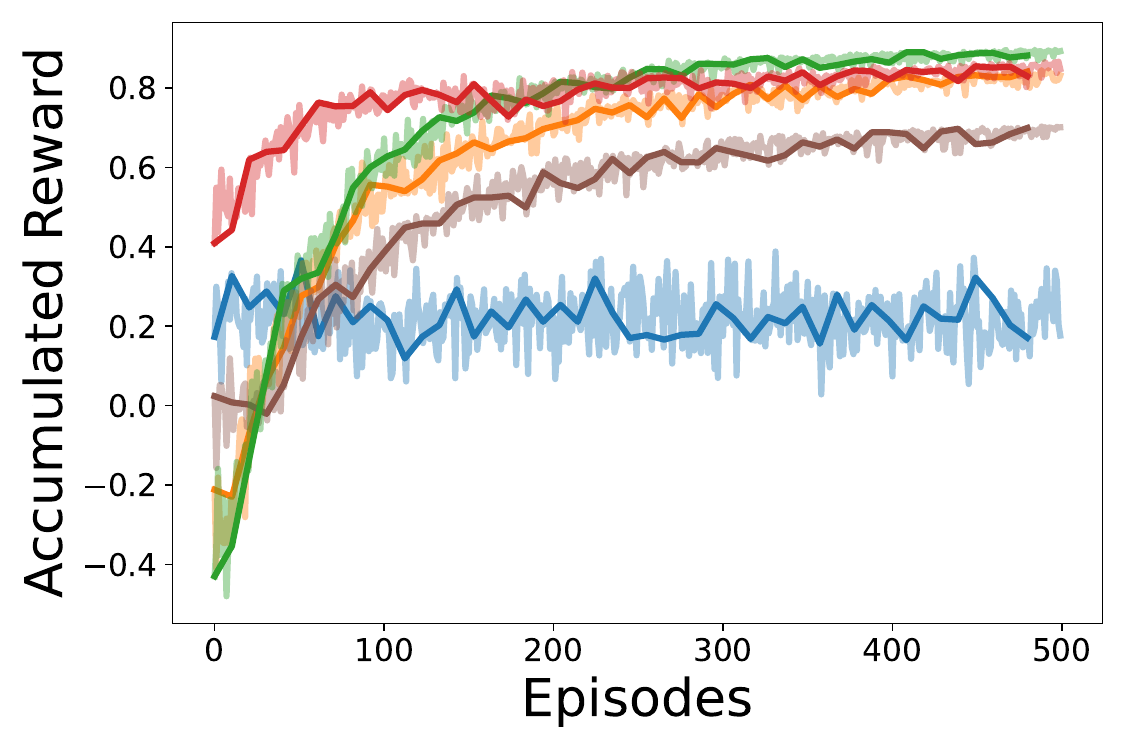} \label{fig:maze_size_5}}
\hfil
\subfloat[$m  = 6$]{\includegraphics[width=0.32\textwidth]{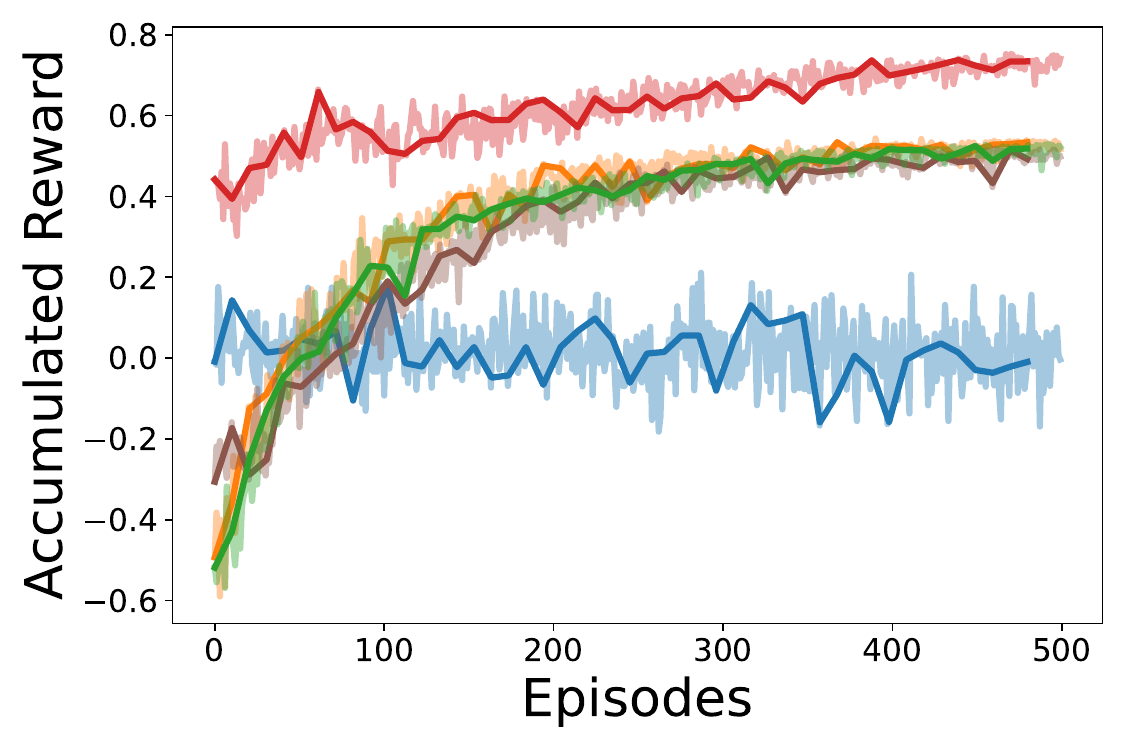} \label{fig:maze_size_6}}
\hfil
\subfloat[$m  = 7$]{\includegraphics[width=0.32\textwidth]{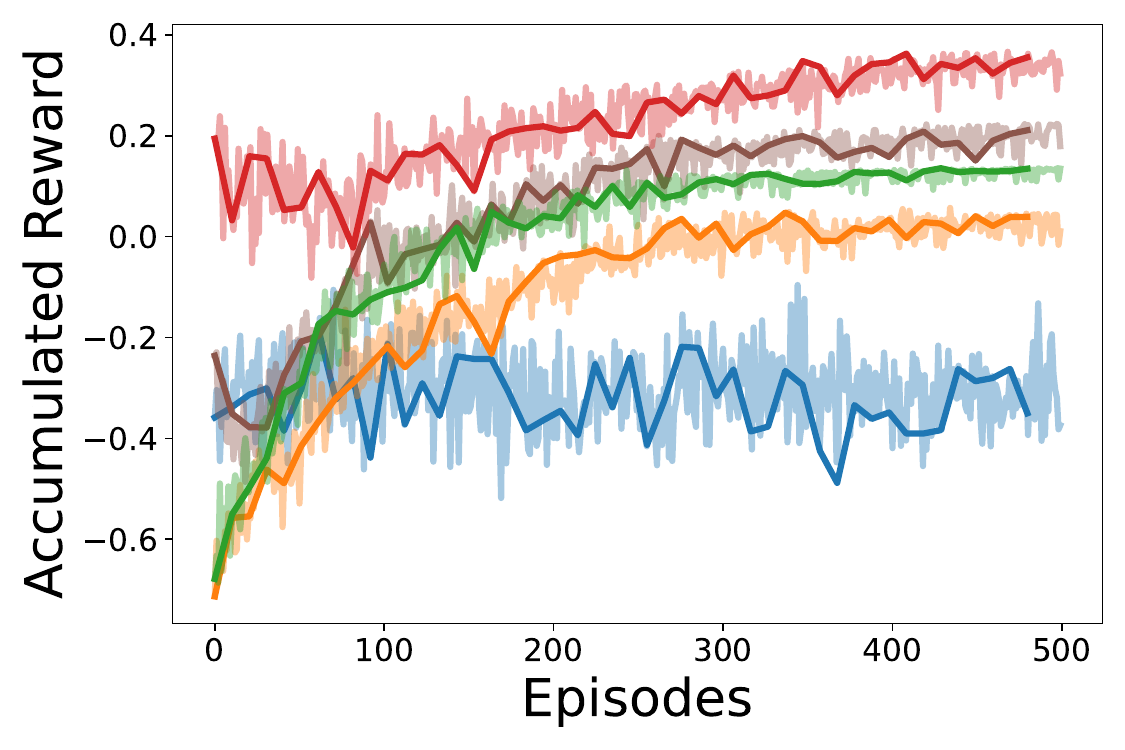} \label{fig:maze_size_7}}
\hfil
\subfloat[$m  = 8$]{\includegraphics[width=0.32\textwidth]{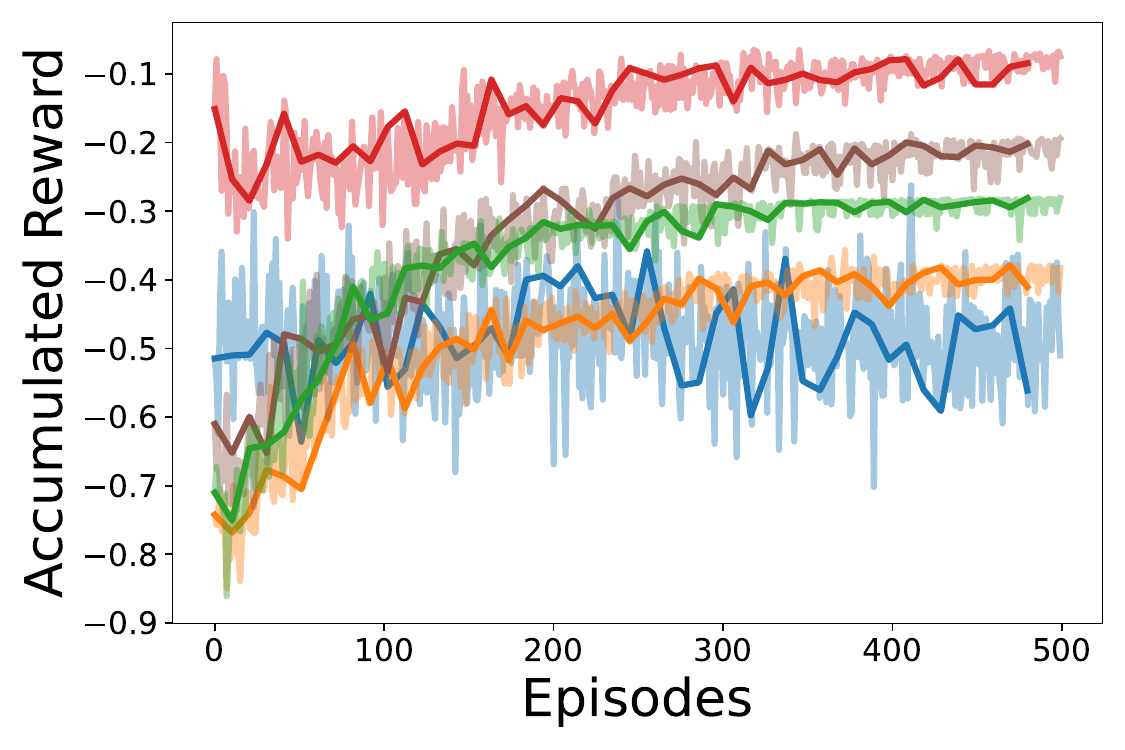} \label{fig:maze_size_8}}
\hfil
\caption{Comparison of training performance under a different maze size ($m$) under DT, PPO, TL PPO, SE PPO, and DT-guided PPO using 500 training episodes in accumulated reward.} \label{fig: maze_train_process}
\end{figure*}

\subsection{Sensitivity Analysis of the Maze Problem}

Figure \ref{fig: maze_train_process} illustrates the dynamic of accumulated rewards as a function of maze size ($m$) ranging from 3 to 8. Each subplot delineates the trajectory of accumulated reward over training episodes, with the x-axis representing the episode count and the y-axis quantifying the reward. Upon examination, we discern several patterns as follows.

First, the DT-guided PPO agent consistently surpasses all other agents (DT, PPO, TL PPPO, and SE PPO) regarding accumulated rewards across all maze sizes. This superior performance suggests that DT-guided PPO leverages the systematic approach of decision theory and the flexible learning capabilities inherent in the PPO's neural network structure. Such an integration gives the agent a robust navigational strategy in the maze, which becomes increasingly advantageous as the maze's complexity escalates. This is particularly crucial in larger mazes where the agent must contend with more intricate challenges. 

Second, notably, as the maze size escalates, the differential in performance between the DT-guided PPO and other PPO agents becomes more pronounced, especially in later episodes. This trend could be attributed to the DT component providing a more effective heuristic in the early stages of exploration, guiding the agent through larger state spaces more efficiently.
The DT's structured approach potentially reduces the search space for the agent by giving zero utility to actions toward obstacles, thereby mitigating the challenges posed by a larger maze's complexity and helping to maintain higher performance levels than the pure PPO agent. 

Third, a general trend observed is the decline in the accumulated reward for all agents with increasing maze size, with the DT agent demonstrating considerable fluctuation when $m \geq 7$. This fluctuation could stem from the amplified complexity and inherent difficulties of larger mazes, where a sole reliance on decision theory might not yield efficient pathfinding consistently. Without the capacity to learn and adapt from interactions with the environment, a purely decision-theoretic model may fail. Conversely, a pure PPO approach may struggle with sparse rewards, as exits become harder to reach and positive rewards become less frequent as the maze expands. 

Despite these challenges, the DT-guided PPO resists the sparse reward problem. This resilience is likely due to the initial guidance provided by the decision theory component, which orients the agent towards more rewarding trajectories, even within the larger mazes. The hybrid model's ability to integrate structured decision-making with adaptive learning proves to be effective in managing the complexities of the maze problem at varying scales.

The DT-guided PPO's consistent outperformance across maze sizes illustrates the value of combining structured decision-making with empirical adaptive neural networks, especially when dealing with problems of increasing size and complexity. The data suggests that this hybrid approach could be a promising direction for developing robust solutions in complex, dynamic environments.

\begin{figure}[h]
\centering
\subfloat{\includegraphics[width=0.49\textwidth]{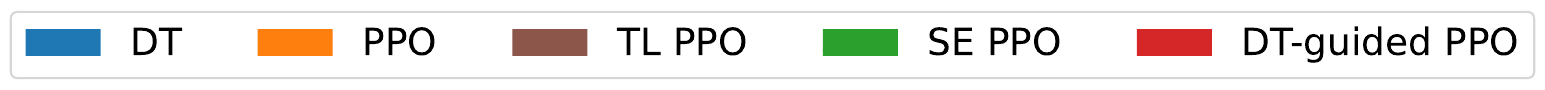}}
\hfil
\vspace{-4mm}

\setcounter{subfigure}{0}
\subfloat{\includegraphics[width=0.48\textwidth]{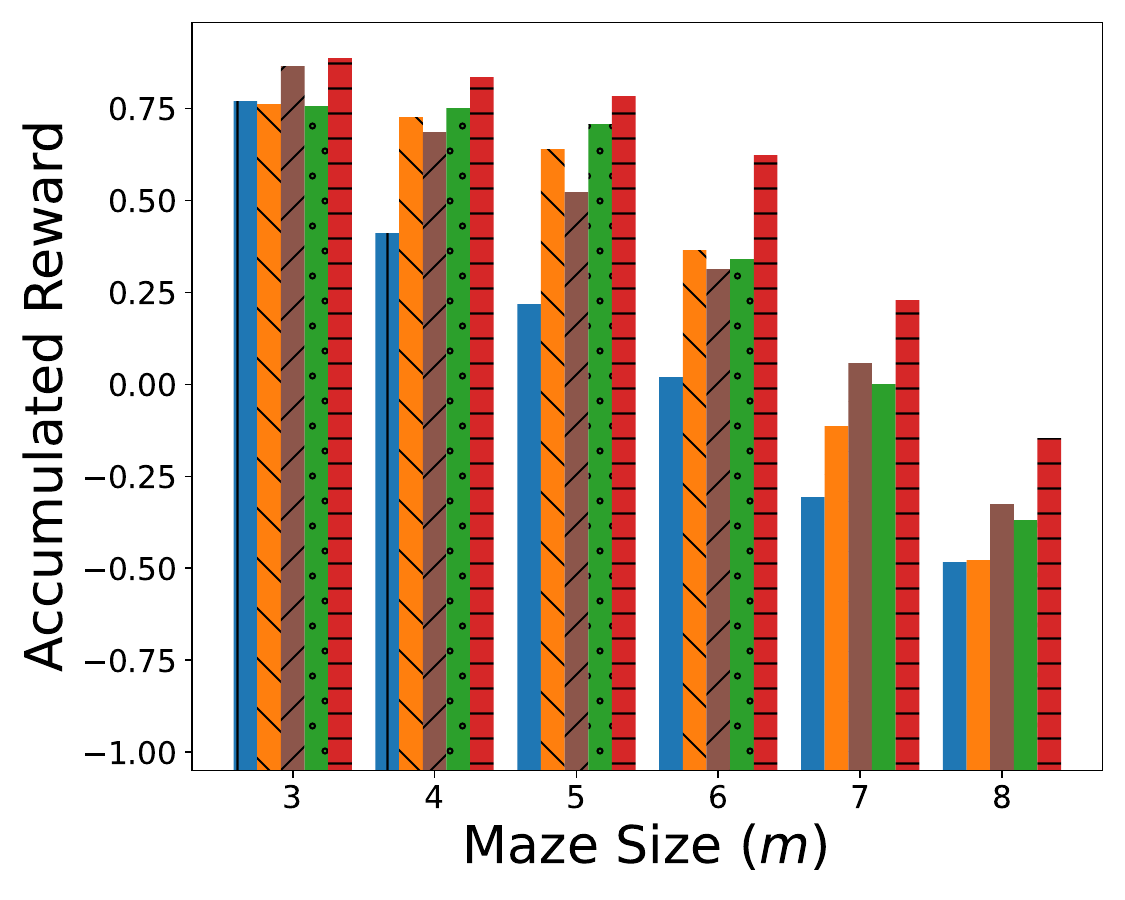} \label{fig: label}}
\hfil
\caption{Comparison of accumulated rewards under varying a maze size ($m$) with the average of over 500 episodes under DT, PPO, TL PPO, SE PPO, and DT-guided PPO.} \label{fig: maze_reward_bar}
\vspace{-3mm}
\end{figure}

Figure \ref{fig: maze_reward_bar} provides a comparative analysis of the performance changes correlated with increasing maze sizes. The x-axis categorizes the maze size, while the y-axis quantifies the accumulated reward for the five distinct agents: DT, PPO, TL PPO, SE PPO, and DT-guided PPO. In sum, we observe the following.

First, a clear trend of decreasing accumulated reward is evident as the maze size expands for all five algorithms. This is also expected in Figure \ref{fig: maze_train_process}, as larger mazes present more complex navigation challenges and potentially increase the likelihood of the agent not finding an exit or taking suboptimal paths, diminishing the overall reward. 

Second, the DT-guided PPO agent shows a consistent advantage in accumulated reward across all maze sizes compared to the DT, PPO, TL PPO, and SE PPO agents. This consistent outperformance indicates the DT-guided PPO's superior capability in balancing structured decision-making with adaptable learning to navigate through mazes of varying complexities. 

Third, the bar chart indicates a point of inflection between maze sizes 6, 7, and 8, where the decline in reward for the DT-guided PPO agent appears less steep than that of the pure PPO agent. This inflection could be interpreted as the point where the heuristic from decision theory guidance begins to significantly impact performance.

\begin{figure}[h]
\centering
\subfloat{\includegraphics[width=0.49\textwidth]{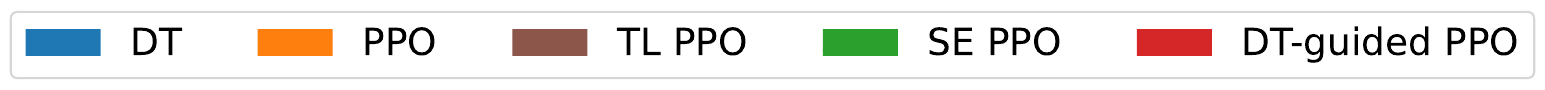}}
\hfil
\vspace{-4mm}
\setcounter{subfigure}{0}
\subfloat{\includegraphics[width=0.48\textwidth]{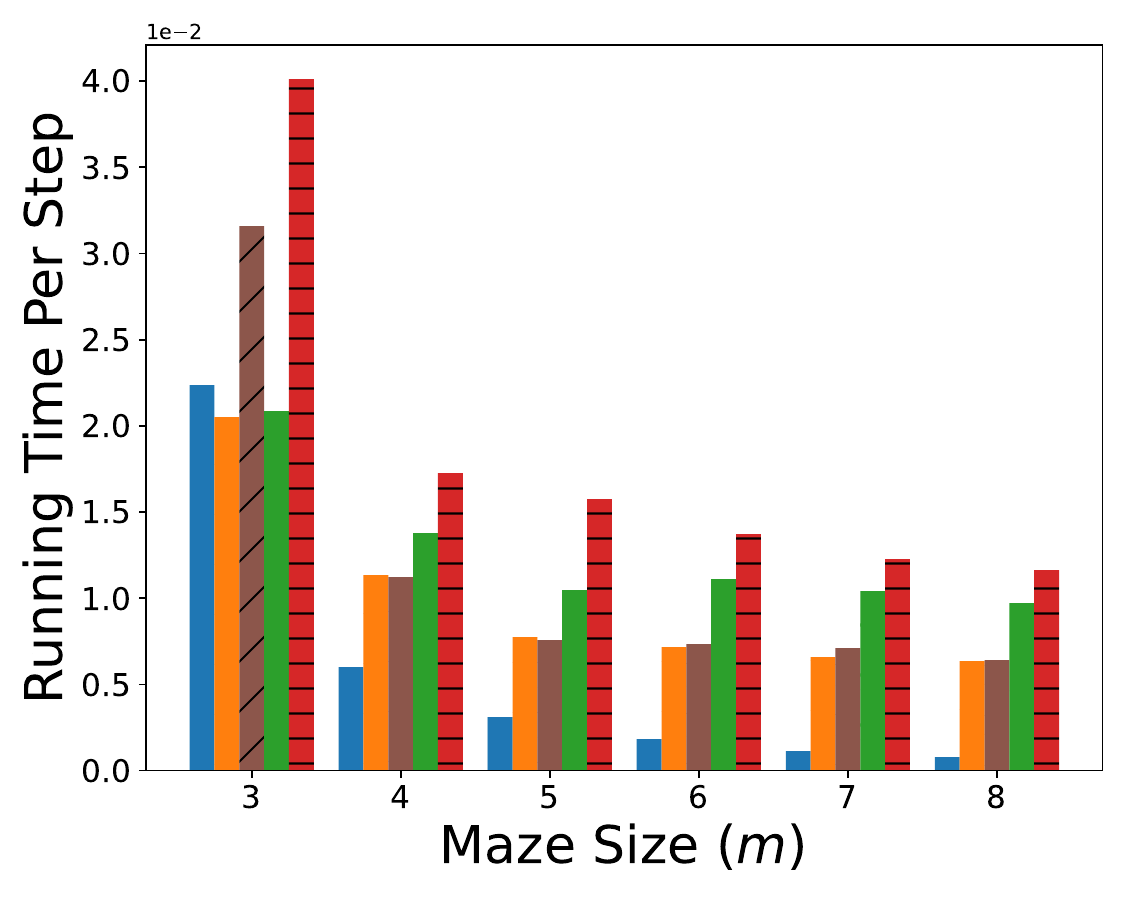} \label{fig: label}}
\hfil
\caption{Comparison of running time per step under varying a maze size ($m$) with the average of over 500 episodes under DT, PPO, TL PPO, SE PPO, and DT-guided PPO.} \label{fig: maze_run_time_bar}
\end{figure}

Figure \ref{fig: maze_run_time_bar} shows the running time per step for each agent as the maze size ($m$) changes. The $y$-axis measures the average time taken per step. Due to the variation in the number of steps per episode, we divide the running time by the total number of steps to obtain a normalized value for fair comparison.  From Figure \ref{fig: maze_run_time_bar}, we observed that the DT-guided PPO algorithm incurs a higher running time per step than other algorithms. This is attributable to the DT-guided PPO's dual computational process, which integrates the decision-theoretic framework with a reinforcement learning component. The additional computation involves not only the neural network's forward and backward propagations but also the evaluation of a utility function, which collectively increases the time complexity per step.

\section{Conclusion}

This paper presented Decision Theory-guided Deep Reinforcement Learning (DT-guided DRL), an innovative approach designed to mitigate the cold start problem prevalent in DRL applications. By integrating decision theory, our proposed DT-guided DRL framework aims to enhance both the initial performance and the robustness of DRL agents, facilitating a more efficient and reliable convergence during learning tasks. Key findings from our experimental evaluations, set within the cart pole and maze problem contexts, underscore the effectiveness of DT-guided DRL. Notably, DT-guided DRL agents demonstrated superior performance in terms of early-stage rewards and learning acceleration compared to conventional DRL agents. Moreover, these agents exhibited remarkable robustness against the increased complexity in larger problem domains, particularly evidenced in the maze problem.

Specifically, our \textbf{key findings} include the following: 
\begin{itemize}
\item In the cart pole scenario, DT-guided DRL achieved higher initial rewards and faster convergence to optimal policies compared to existing DRL techniques, like transfer learning, sample efficiency, and imitation learning. This underscores the benefit of decision theory-based heuristics in the early learning phases. 
\item In the maze problem, DT-guided DRL consistently outperformed existing approaches across various maze sizes, showcasing its adaptability and effectiveness in complex environments with sparse rewards. 
\item The integration of decision theory not only provided effective initial guidance for the DRL agents but also contributed to a more structured and informed exploration strategy, particularly in environments with large state spaces and intricate navigational challenges.
\end{itemize}

In conclusion, DT-guided DRL represents a significant step forward in addressing some of its intrinsic challenges by combining the principles from equation-based functions designed based on human (designer) knowledge. While our initial findings are promising, extensive research is required to fully exploit the potential of this novel approach. We anticipate our work will inspire further exploration and innovation at the intersection of decision theory and deep reinforcement learning. We encourage the community to build upon our work; the source code is accessible at \url{https://github.com/Wan-ZL/DT-DRL}.

\section*{Acknowledgment}
This paper presents work whose goal is to advance the field of Machine Learning. There are many potential societal consequences of our work, none of which we feel must be specifically highlighted here.

\newpage
\bibliography{ref}
\bibliographystyle{icml2024}

\end{document}